\documentclass[10pt,twocolumn,letterpaper]{article}

\usepackage[pagenumbers]{cvpr}      

\usepackage{graphicx}
\usepackage{amsmath}
\usepackage{amssymb}
\usepackage{amsthm}
\usepackage{booktabs}
\usepackage{multirow} 
\definecolor{cvprblue}{rgb}{0.21,0.49,0.74}
\usepackage[pagebackref,breaklinks,colorlinks,allcolors=cvprblue]{hyperref}
\newtheorem{remark}{Remark}

\usepackage{algorithm}
\usepackage{algorithmic}

\title{Meta-Task: A Method-Agnostic Framework for Learning to Regularize in Few-Shot Learning}

\author {
    Mohammad~Rostami\\
     Rowan University\\
    {\tt\small rostami23@rowan.edu}
    \and
    Atik Faysal\\
    Rowan University\\
    {\tt\small faysal24@rowan.edu}
    \and
    Huaxia Wang\\
    Rowan University\\
    {\tt\small huaxia.wang@rowan.edu}
    \and
    Avimanyu~Sahoo\\
    University of Alabama in Huntsville\\
    {\tt\small avimanyu.sahoo@uah.edu}\\
}

\begin{document}
\maketitle
\begin{abstract}
Overfitting is a significant challenge in Few-Shot Learning (FSL), where models trained on small, variable datasets tend to memorize rather than generalize to unseen tasks. Regularization is crucial in FSL to prevent overfitting and enhance generalization performance. To address this issue, we introduce {{Meta-Task}}, a novel, method-agnostic framework that leverages both labeled and unlabeled data to enhance generalization through auxiliary tasks for regularization. Specifically, Meta-Task introduces a Task-Decoder, which is a simple example of the broader framework, that refines hidden representations by reconstructing input images from embeddings, effectively mitigating overfitting.

Our framework’s method-agnostic design ensures its broad applicability across various FSL settings. We validate Meta-Task’s effectiveness on standard benchmarks, including Mini-ImageNet, Tiered-ImageNet, and FC100, where it consistently {\color{black} improves} existing {\color{black} state-of-the-art} meta-learning techniques, demonstrating superior performance, faster convergence, reduced generalization error, and lower variance—all without extensive hyperparameter tuning. These results underline Meta-Task’s practical applicability and efficiency in real-world, resource-constrained scenarios.
\end{abstract}
\begin{figure}
  \centering
   \includegraphics[width=1.1\linewidth]{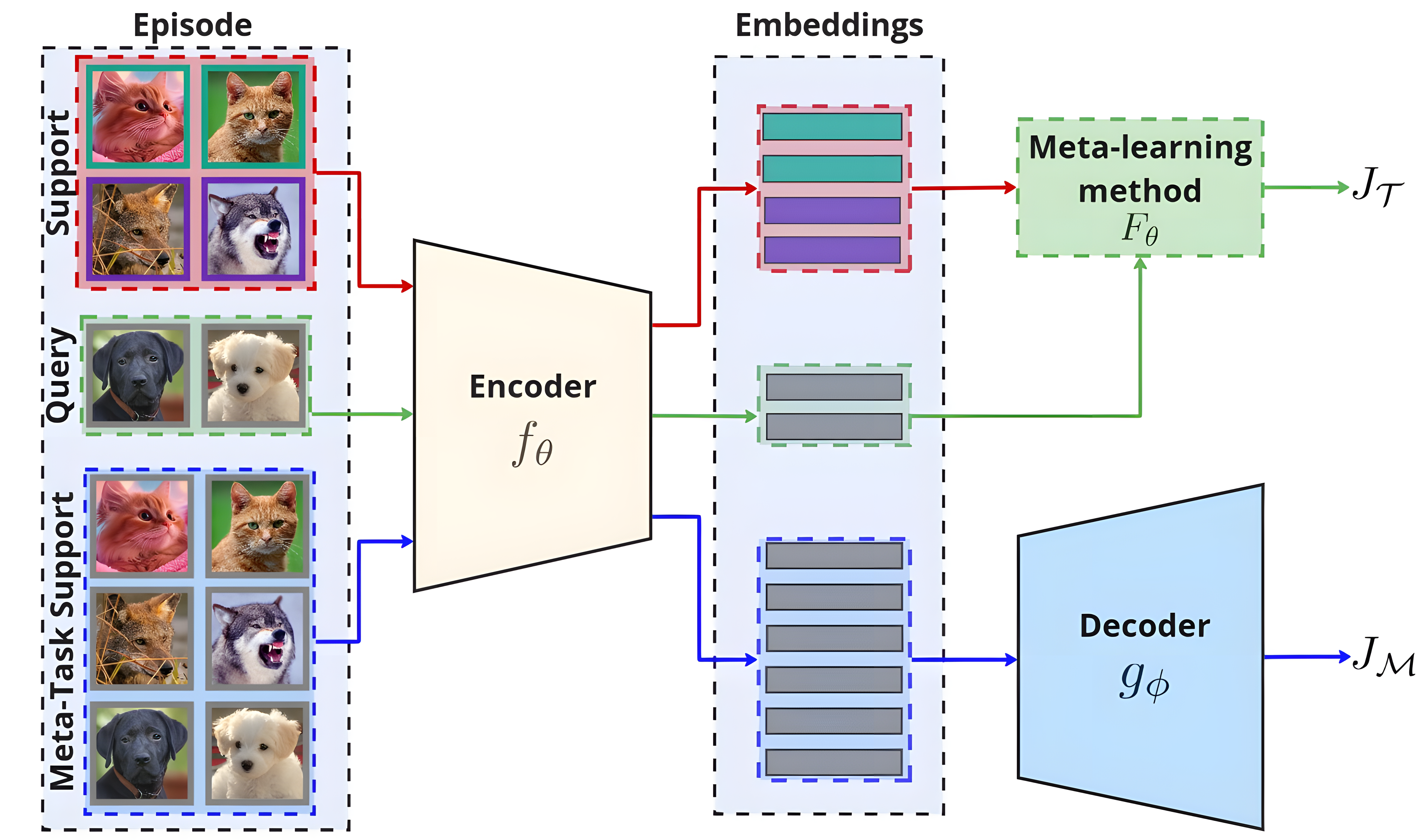}   \caption{Detailed architecture of the Task-Decoder within the Meta-Task framework. The Meta-Task support set consists of unlabeled copies of support and query samples, which are fed into an encoder \( f_{\theta} \) to produce feature embeddings. These embeddings serve two key functions: (1) they are passed to a decoder \( g_{\phi} \) to reconstruct the input images, aiding in the learning of rich, meaningful representations, and (2) they are used to generate the meta-learning method ($F_{\theta}$) for query image classification. The model optimizes two objectives: classification loss (\( J_{\mathcal{T}} \)) and reconstruction loss (\( J_{\mathcal{M}} \)), each with a dedicated backpropagation step. The overall training loss is \( J_{\mathcal{T}} + \lambda J_{\mathcal{M}} \), balancing both classification accuracy and representation learning. For a broader application, the decoder \( g_{\phi} \) could be replaced by one or more neural networks that solve various unsupervised tasks.}

   \label{fig:1}
\end{figure}    
\section{Introduction}
\label{sec:intro}

{\color{black}{Few-Shot Learning (FSL) has emerged as a promising solution to the problem of data scarcity in machine learning, particularly in computer vision, medical applications, and personalized health systems\cite{nayem2023few, PFM_Ali, Event_Ali}. By leveraging meta-learning, FSL enables models to learn how to learn\cite{learn2learn} with the primary objective of extracting meaningful, transferable representations from sparse datasets, reducing the need for extensive labeled data, and significantly lowering data collection and annotation costs~\cite{metasgd, support, metagan}.

Advances in meta-learning have led to techniques such as Matching Networks~\cite{matching}, Relation Networks~\cite{relation}, Model-Agnostic Meta-Learning (MAML)\cite{maml}, Prototypical Networks\cite{proto}, and Reptile~\cite{reptile}. Despite performance gains, FSL models remain prone to overfitting, limiting generalization.

Regularization plays a critical role in mitigating overfitting and enhancing generalization in FSL. Conventional methods such as weight decay\cite{krogh1991simple}, dropout\cite{hinton2012improving}, and noise injection\cite{achille2018information, alemi2016deep, tishby2015deep} fail to address FSL-specific challenges. Meta-learning requires both adaptation-generalization and meta-generalization\cite{imaml, mrmaml, pan2021meta, wang2023improving, yao2021meta, yao2021improving, rajendran2020meta}, yet existing techniques remain static, inflexible, and reliant on manually designed loss functions or auxiliary tasks that do not adapt to the unique structure of each task. This limitation motivates the need for a more dynamic, learnable regularization mechanism.}}

{\color{black}{To address this limitation, we introduce Meta-Task, a method-agnostic framework that reformulates regularization as a learnable task within the FSL optimization process. Meta-Task dynamically learns task-specific regularization strategies through a parameterized neural layer $\phi$. This adaptive mechanism enables our framework to seamlessly integrate with existing FSL pipelines, ensuring flexibility across different task distributions and learning paradigms. Meta-Task remains compatible with existing auxiliary tasks and regularization losses, leveraging prior techniques while introducing a dynamic, learnable component. This enhances overfitting mitigation without extensive hyperparameter tuning or task-specific modifications, making it scalable and practical for real-world FSL applications.}}

{\color{black}{To operationalize Meta-Task regularization, we propose using an autoencoder-based Task-Decoder (Figure \ref{fig:1}). The Task-Decoder serves as a flexible and effective regularizer, refining task-specific representations by constraining the embedding vector to reconstruct the original input accurately. We choose autoencoders because they are data-agnostic, easy to implement, architecturally flexible, and allow tunable parameterization for optimization. This adaptability makes them well-suited for various FSL applications, as they can seamlessly integrate into different model architectures.

Our method leverages the latent variable $z$, inherently present in most FSL settings due to the N-Way, K-Shot paradigm. The standard inner-outer loop optimization structure ensures the existence of an intermediate latent representation within the support-query split or gradient-based adaptation process. This guarantees compatibility across various FSL architectures, allowing Meta-Task to enhance regularization effectiveness dynamically.}}

The main contributions of this paper are as follows. (1) Meta-Task Framework: We propose Meta-Task, a novel, method-agnostic framework that introduces auxiliary tasks to regularize individual tasks in FSL. (2) Task-Decoder (Figure \ref{fig:1}): We introduce the Task-Decoder, utilizing autoencoders as a new regularization task for meta-learners, leveraging unsupervised learning to refine hidden representations. (3) Improved Performance: Our method demonstrates faster convergence, higher accuracy, and better generalization than existing techniques, all without extensive hyperparameter tuning. (4) Versatility and Applicability: The framework is method-agnostic and can be integrated with various FSL settings, making it suitable for real-world, resource-constrained scenarios.

The remainder of the paper is organized as follows: Section \ref{sec:back_ground} reviews related work to contextualize the problem. Section \ref{sec:approach} details our proposed approach, and Section \ref{sec:results} presents experimental results. Finally, conclusions are drawn in Section \ref{sec:conclusion}.

\section{Related Work}\label{sec:back_ground}

This section briefly reviews the relevant literature on meta-learning and regularization issues. Regularization techniques have been developed to improve the generalization ability of meta-learning models. These techniques can be broadly classified into four specialized regularization techniques for meta-learning. 1) Explicit Regularization (ER) \cite{imaml,mrmaml,pan2021meta,wang2023improving}, 2) Episode Augmentation Regularization (EAR) \cite{yamaguchi2023regularizing, shu2023dac}, 3) Task Augmentation Regularization (TAR) \cite{yao2021meta,yao2021improving,rajendran2020meta}, 4) Adversarial Task Upsampling (ATU) \cite{wu2022adversarial}. 

The ER methods impose explicit regularization terms on the meta-learning update, such as Meta-Learning with Implicit Gradients (iMAML) \cite{imaml}, Meta-Learning without Memorization (MR-MAML) \cite{mrmaml}, and their combinations or extensions \cite{pan2021meta, wang2023improving}. These methods directly constrain the model optimization process, aiming to improve the generalization performance of meta-learning models. For instance, minimax-meta regularization \cite{wang2023improving}
combines two types of regularizations at the inner and outer levels to enhance the generalization performance of bi-level meta-learning.

The EAR approach controls the meta-training by modifying individual tasks through noise or mixup \cite{yamaguchi2023regularizing, shu2023dac}. By expanding the amount of training data available for sampling, data augmentation supports various aspects of the meta-learning pipeline. It can be applied to support data in the inner loop of fine-tuning and enlarge the pool of evaluation data to be sampled during training. 

TAR methods generate new tasks by interpolating between or augmenting existing tasks \cite{yao2021meta, rajendran2020meta}, aiming to expand the task distribution for more robust generalization. For instance, methods such as MetaMix \cite{chen2021metamix}, and Channel Shuffle \cite{yao2021improving} have been proposed to increase the dependence of target predictions on the support set and provide additional knowledge to optimize model initialization. ProtoDiff \cite{du2023protodiff} enhances prototypical networks through a task-guided diffusion process; by generating synthetic features conditioned on the task, ProtoDiff \cite{du2023protodiff} augments the support set and improves the robustness of prototypes, significantly boosting classification performance. Additionally, the ATU approach \cite{wu2022adversarial} seeks to generate tasks that match the true task distribution by maximizing an adversarial loss, thereby improving meta-testing performance and the quality of upsampled tasks. These task augmentation techniques are designed to enhance the generalization capability of meta-learning models by effectively expanding the task distribution and mitigating overfitting during meta-training.


Recent works like Meta-Baseline \cite{chen2021meta} have advanced few-shot learning by bridging transfer learning and meta-learning through pre-training feature extractors on base classes before few-shot fine-tuning, demonstrating that high-quality feature representations can sometimes eliminate the need for complex meta-learning algorithms. However, in contrast to Meta-Baseline's reliance on large-scale pre-training and ProtoDiff's \cite{du2023protodiff} computationally intensive task-guided diffusion models, our method learns robust embeddings during meta-training itself, making it more feasible for resource-constrained environments while remaining computationally efficient and method-agnostic.

In contrast to these methods, our proposed framework aims to regularize meta-learning tasks by introducing Meta-Tasks—auxiliary tasks designed to refine hidden representations without relying on extensive pre-training or synthetic data generation. Table \ref{tab:meta_learning_comparison} summarizes the key differences between various methods and our approach. We integrate an unsupervised autoencoder, specifically the Task-Decoder, directly within the meta-learning framework to provide continuous, unsupervised regularization during training. This helps prevent overfitting and improves generalization in an end-to-end manner.


By treating regularization as a learnable task, our framework enhances adaptability and generalization capabilities in a lightweight manner. It is suitable for a broader range of real-world applications where computational resources and labeled data are limited.

{\color{black}{Meta-Task distinguishes itself from prior FSL approaches by introducing a learnable, task-specific regularization strategy that dynamically adapts to each task without modifying the underlying FSL architecture. Unlike LEO and iMAML, which focus on optimizing meta-initialization and gradient regularization, Meta-Task directly integrates into the optimization process through a parameterized neural layer $\phi$, learning task-adaptive constraints rather than enforcing fixed optimization heuristics. Unlike Wertheimer et al. (2021) and Nag et al. (2023), which utilize reconstruction-based self-supervision, our method does not rely on reconstruction loss. Furthermore, unlike Gidaris et al. (2019), which introduces auxiliary tasks, Meta-Task requires neural layer $\phi$, instead only relying on better updates using self-supervised or semi-supervised tasks.}}

Our comprehensive experiments focus on fundamental meta-learning models (Prototypical Networks, MAML, MetaOptNet, {\color{black} and P$>$M$>$F~\cite{pmf}}) that represent diverse approaches in the field. While some baseline comparisons were conducted with adjusted experimental settings (reduced batch sizes, modified query shots, etc.) due to computational constraints, our results consistently demonstrate the effectiveness of our Meta-Task framework across different meta-learning paradigms, validating its method-agnostic nature and broad applicability.
\begin{table}[ht]
\centering
\resizebox*{0.5\textwidth}{!}{
\begin{tabular}{lccccc}
\toprule
\textbf{Method} & \textbf{Category} & \textbf{Method-Agnostic} & \textbf{Complexity} \\
\midrule
\textbf{iMAML} \cite{imaml} & ER & No & High
\\
\textbf{MR-MAML} \cite{mrmaml} & ER & No & High \\ 
\textbf{Minimax-Meta} \cite{wang2023improving} & ER & No & Low \\
\textbf{DAC-MR} \cite{shu2023dac} & EAR & Yes & Medium \\
\textbf{MetaMix} \cite{chen2021metamix} & EAR & Yes & Medium \\
\textbf{ProtoDiff} \cite{du2023protodiff} & TAR & No & High \\
\textbf{ATU} \cite{wu2022adversarial} & TAR & Yes & High \\
\textbf{LEO} \cite{leo} & ER & No & High \\
\textbf{Task-Decoder (Ours)} & \textbf{MTR} & \textbf{Yes} & \textbf{Medium} \\
\bottomrule
\end{tabular}
}
\caption{Comparison of meta-learning regularization methods. The table compares various meta-learning regularization techniques by category, method-agnostic capability, and computational complexity. Categories include Explicit Regularization (ER), Episode Augmentation Regularization (EAR), Task Augmentation Regularization (TAR), and our proposed Meta-Task Regularization (MTR).}
\label{tab:meta_learning_comparison}

\end{table}







\section{Meta-Task Framework}\label{sec:approach}
In this section, we introduce our proposed Meta-Task Framework, a novel approach that enhances generalization by integrating learnable Meta-Tasks as a form of regularization. By treating regularization itself as a learnable task, our framework encourages the model to develop richer and more robust feature representations. The framework is method-agnostic and can be seamlessly integrated with existing FSL models, such as Prototypical Networks and MAML. In the following subsections, we detail the components of the framework, including the Task-Decoder, the problem formulation, and the theoretical justification for its effectiveness.

Our framework extends MAML \cite{maml} by introducing regularization as a \textit{learnable Meta-Task}, enabling unsupervised task integration during meta-training without requiring additional data. The following subsections detail the implementation and theoretical foundations of our contributions:

\begin{itemize}
\item \textbf{Task-Decoder:} We introduce an unsupervised autoencoder architecture that implements our learnable regularization concept by reconstructing input images from their embeddings, proving its effectiveness in enriching feature representations.
\item \textbf{Problem Formulation:} We present the mathematical framework that enables the integration of unsupervised auxiliary tasks with existing FSL methods, demonstrating how our approach remains method-agnostic through separable loss functions.
\item \textbf{Theoretical Analysis:} We provide theoretical justification for our framework's convergence properties and establish how the joint optimization of primary and Meta-Tasks enhances generalization capabilities.
\end{itemize}

\subsection{Task-Decoder}

At the core of our Meta-Task Framework is the Task-Decoder (Figure \ref{fig:1}), an unsupervised learning component designed to enhance the embedding space by reconstructing input images from their embeddings. The Task-Decoder employs an autoencoder architecture, where the encoder \( f_{\theta} \) maps the input images to a latent embedding space, and the decoder \( g_{\phi} \) reconstructs the images from these embeddings.

By encouraging the embeddings to retain sufficient information for accurate reconstruction, the Task-Decoder prevents the encoder from focusing solely on discriminative features that may not generalize well. This leads to richer and more robust feature representations, ultimately improving the model's performance on unseen tasks.

The Task-Decoder operates in parallel with the primary few-shot learning model, processing the same inputs and contributing to the overall learning objective through its reconstruction loss.

\subsection{Problem Formulation}

Let the Meta-Task $\mathcal{M}$ denote the Task-Decoder, and our meta-learning method $F$ be Prototypical Networks. We define the encoder network \( f_{\theta} \) parameterized by \( \theta \), and \( g_{\phi} \) represent the decoder network parameterized by \( \phi \). 

For a given query sample \( x\in\mathcal{Q} \) and its true class \( y \) belonging to task $\mathcal{T}$, the Prototypical Network $F_{\theta}$ computes class prototypes using support set $\mathcal{S}$ and classifies $x$ based on the closest prototype to the embedding $f_{\theta}$.

\noindent
\textbf{Primary Task Loss:} The probability of \( x \) belonging to class \( n \) is:

\begin{equation} p_{\theta}(y={n} \mid x) = \frac{\exp\left(-d\left(f_{\theta}(x), c_{{n}}\right)\right)}{\sum_{k=1}^{N} \exp\left(-d\left(f_{\theta}(x), c_k\right)\right)}, \end{equation}
where $N$ is the number of tasks in the episode, \( c_n \) is the prototype of class \( n \) and \( d(\cdot, \cdot) \) denotes the Euclidean distance. The classification loss is then:

\begin{equation}
J_{\mathcal{T}}(x, \theta) = -\log p_{\theta}(y = n \mid x).
\end{equation}

\noindent
\textbf{Meta-Task Loss:} The Task-Decoder aims to reconstruct \( x \) from \( f_{\theta} \), with the reconstruction loss defined as:

\begin{equation}
J_{\mathcal{M}}(x, \theta, \phi) = \left\| g_{\phi}(f_{\theta}(x)) - x \right\|^2.
\end{equation}

\noindent
\textbf{Combined Loss Function:} The total loss for a sample \( x \) is a weighted sum of the primary task loss and the Meta-Task loss:

\begin{equation}
J(x, \theta) = J_{\mathcal{T}}(x, \theta) + \lambda J_{\mathcal{M}}(x, \theta, \phi),
\end{equation}
where \( \lambda \) is a hyperparameter balancing the two losses or the learning rate for the Meta-Task update.

\noindent
\textbf{Optimization:} The model parameters \( \theta \) and \( \phi \) are optimized jointly over all tasks and samples. For a given task $\mathcal{T}$, the overall loss is calculated by summing the combined losses over all query samples:

\begin{equation}\label{eq:task_loss}
J_{\mathcal{T}}(\theta) = \sum_{x \in \mathcal{T}} J_{\mathcal{T}}(x, \theta). 
\end{equation}

During meta-training, we sample a batch of tasks $\{\mathcal{T}_i\}^L_{i=1}$. The total loss across all tasks in an episode is:

\begin{equation}\label{eq:overall_loss_separate}
J(\theta) = \sum_{i=1}^{L} \left(J_{\mathcal{T}_i}(\theta) + \lambda J_{{\mathcal{M}_i}}(\theta, \phi)\right).
\end{equation}

By minimizing this combined loss, the model learns embeddings that are both discriminative for classification and informative for reconstruction, enhancing generalization across tasks. The full algorithm of the Task-Decoder integration is provided in Algorithm~\ref{alg:task_decoder}.

We define tasks and Meta-Tasks as separable if their loss functions can be computed independently and summed without including interaction terms. This structure allows us to optimize each component separately while still contributing to the overall objective. If the regularization term is separable (as in our case with the autoencoder), we can simplify \eqref{eq:overall_loss_separate} as
\begin{equation}\label{eq:9}
J(\theta) = \sum_{i=1}^{L} J_{\mathcal{T}_i}(\theta) +\lambda\sum_{i=1}^{L}J_{{\mathcal{M}_i}}(\theta,\phi).
\end{equation}

If each task may have multiple or no Meta-Tasks attached, the overall loss function generalizes to:
\begin{equation}\label{eq:10}
J(\theta) = \sum_{i=1}^{L} J_{\mathcal{T}_i}(\theta) +\lambda\sum_{r=1}^{R}J_{{\mathcal{M}_r}}(\theta,\phi),
\end{equation}
where $R$ is the total number of regularizers applied to the model.

{\color{black} \subsection{Justification}

Our Meta-Task framework provides unique theoretical guarantees, distinguishing it from existing regularization approaches. In Equations \eqref{eq:9} and \eqref{eq:10}, we demonstrated that incorporating auxiliary tasks into training effectively reformulates the FSL paradigm to a new FSL setting with a combined distribution of primary tasks and Meta-Tasks, denoted as 
$\mathcal{T}'$. This reformulation allows the overall training objective to be expressed as a sum over $L+R$ tasks, preserving the structure necessary for standard FSL optimization techniques.

When using MAML as the FSL method, based on convergence guarantees proven in literature~\cite{wang2022global}, our approach converges to an optimal set of parameters for $\mathcal{T}'$. Since the Meta-Tasks are carefully designed to enhance generalization, this optimal parameter set enables the model to generalize effectively across both primary tasks and Meta-Tasks. The convergence of $\phi$ to its global optima further ensures that the learned embeddings retain the essential information necessary for both classification and reconstruction. This joint optimization ($\theta$) constructs a globally optimal representation space, improving overall model performance on unseen tasks by reducing generalization error.

A critical advantage of Meta-Task is the separability of the primary and auxiliary tasks. Since their loss functions can be computed independently, optimization can proceed efficiently, with each task contributing to the overall objective. This design simplifies the optimization process and supports convergence to a global optimum, as the auxiliary tasks act as an implicit regularizer, reducing overfitting by penalizing non-generalizable embeddings.

A rigorous mathematical proof is more challenging for FSL methods beyond MAML. However, as all FSL methods aim to optimize parameters that generalize across task distributions with limited examples, it is reasonable to infer that our approach improves generalization capabilities, as demonstrated in the results section. In the supplementary material, we provide additional experiments, including visualizations of reconstructed features, to empirically validate how Meta-Task enhance embedding quality and task generalization.}

\begin{remark}{The proposed regularization approach is method-agnostic and can be applied to any FSL learning method.} \end{remark}
\begin{remark}{Equations \eqref{eq:9}, \eqref{eq:10} hold only if tasks and Meta-Tasks are separable or independent.}
\end{remark}
{\color{black}\begin{remark}{The use of autoencoders is because autoencoders can be used with any type of data, are easy to use, and are flexible in architecture.}
\end{remark}}

\begin{algorithm}[t]
\caption{Meta-Tasks: Task-Decoder}
\label{alg:task_decoder}
\begin{algorithmic}[1]
\REQUIRE $p(\mathcal{T})$: distribution over tasks; 
\REQUIRE $f_{\theta}$: embedding network;
\REQUIRE $g_{\phi}$: decoder network; 
\REQUIRE $\alpha$, $\lambda$: learning rates;
\REQUIRE $F$: meta-learning method (e.g., Prototypical Networks);
\REQUIRE $J_{\mathcal{T}}(x, \theta)$: classification loss;
\\$J_{\mathcal{M}}(x, \theta, \phi) = \| g_{\phi}(f_{\theta}(x)) - x \|^2$: reconstruction loss

\STATE Randomly initialize $\theta$ and $\phi$
\WHILE{not done}
    \STATE Sample support set: $\mathcal{S} \sim p(\mathcal{T})$
    \STATE Sample query set: $\mathcal{Q} \sim p(\mathcal{T})$
    \STATE Compute reconstruction loss:\\
    $~~~~~~~~~~~~~~~~~{J}_{\mathcal{M}}(\theta, \phi) \leftarrow \frac{1}{|\mathcal{S} \cup \mathcal{Q}|} \sum_{x \in \mathcal{S} \cup \mathcal{Q}} J_{\mathcal{M}}(x, \theta, \phi)$
    \STATE Use the support set $\mathcal{S}$ and embedding network $f_{\theta}$ to get meta-learning method $F_{\theta}$.
    \STATE Compute classification loss on query set:\\
    $
    ~~~~~~~~~~~~~~~~~~~~{J}_{\mathcal{T}}(\theta) \leftarrow \frac{1}{|\mathcal{Q}|} \sum_{x \in \mathcal{Q}} J_{\mathcal{T}}(x, F_{\theta})
    $
    \STATE Update embedding network parameters:\\
    $
    ~~~~~~~~~~~~~~~~~~~~\theta \leftarrow \theta - \alpha \nabla_{\theta} \left( {J}_{\mathcal{T}}(\theta) + \lambda {J}_{\mathcal{M}}(\theta, \phi)) \right)
    $
    \STATE Update decoder parameters:\\
    $
    ~~~~~~~~~~~~~~~~~~~~~~~~~~\phi \leftarrow \phi - \lambda \nabla_{\phi} {J}_{\mathcal{M}}(\theta, \phi)
    $
\ENDWHILE
\end{algorithmic}
\end{algorithm}

\section{Experiments and Results}\label{sec:results}
In this section, we conduct comprehensive experiments to validate the effectiveness of our proposed {Meta-Task} framework in enhancing generalization and mitigating overfitting in few-shot learning. Our objectives are to: 
\begin{enumerate} 
\item Demonstrate that incorporating Meta-Tasks as auxiliary regularization improves model generalization to both seen and unseen tasks.
\item Validate that leveraging both labeled and unlabeled data through an unsupervised autoencoder refines hidden representations.
\item Showcase the method-agnostic applicability of {Meta-Task} across various FSL settings and its practical efficiency in resource-constrained scenarios. 
\end{enumerate}

We begin by outlining the experimental setup, including datasets, model architectures, and implementation specifics. Following this, we present our results along with a comprehensive analysis of our findings, focusing on the impact of the {Meta-Task} framework on performance, convergence speed, and variance reduction.

\subsection{Datasets}

We evaluated the proposed {Meta-Task} framework on three standard benchmarks commonly used in few-shot learning:

\noindent
\textbf{MiniImageNet}~\cite{mini-imagenet}: This dataset consists of 100 classes selected from ImageNet, with 600 images per class resized to $84 \times 84$ pixels. It is divided into 64 training, 16 validation, and 20 test classes, ensuring that test classes are entirely disjoint from training classes for robust evaluation of generalization capabilities.

\noindent
\textbf{TieredImageNet}~\cite{ren2018meta}: A larger and more challenging dataset comprising 608 classes organized into 34 high-level categories, providing semantic diversity. It includes 351 training, 97 validation, and 160 test classes, with images resized to $84 \times 84$ pixels. Its hierarchical structure makes it suitable for evaluating models' ability to generalize across different levels of abstraction.

\noindent
\textbf{FC100}~\cite{oreshkin2018tadam2}: Derived from the CIFAR-100 dataset, FC100 contains 100 classes, each with 600 images resized to $32 \times 32$ pixels. The dataset is split into 60 training, 20 validation, and 20 test classes. It presents unique challenges due to low-resolution images and high intra-class variability, making it an excellent benchmark for assessing model robustness.

These datasets offer diverse and challenging scenarios for few-shot learning, enabling us to thoroughly evaluate the generalization capabilities of the {Meta-Task} framework.

\subsection{Methods}

To demonstrate the method-agnostic nature of our approach, we integrated the {Meta-Task} framework with four few-shot learning models:

\noindent
\textbf{Prototypical Network (PN)} \cite{proto}: Utilizes a ResNet-50 \cite{resnet} backbone for feature extraction. It computes class prototypes in the embedding space, classifying query samples based on their distances to these prototypes.

\noindent
\textbf{MAML}~\cite{maml}: Employs a four-layer ConvNet architecture aiming to find initial model parameters that can be quickly adapted to new tasks with minimal gradient updates.

\noindent
\textbf{MetaOptNet}~\cite{lee2019meta}: Uses a ResNet-12 backbone and incorporates differentiable convex optimization layers for efficient classification, seamlessly integrating with our {Meta-Task} framework.

{\color{black} \noindent
\textbf{P$>$M$>$F} \cite{pmf}: Uses DINO\cite{dino} framework as its backbone. Like PN, it computes class prototypes in the embedding space. It classifies query samples based on their distances to these prototypes but with the added benefit of leveraging pre-trained transformer-based feature extraction.}

In our {Meta-Task} framework, we introduced a Task-Decoder employing an unsupervised autoencoder to reconstruct input images from their embeddings. The decoder mirrors the encoder architecture used in the backbone models. This process leverages both labeled and unlabeled data (support and query sets), refining the hidden representations and mitigating overfitting.

For the autoencoder architecture, we experimented with different configurations to assess the impact of the Task-Decoder's depth on performance. Specifically, we used both shallow and deep autoencoders, with parameters ranging from 75K to 250K. The shallow autoencoder consists of two convolutional layers, while the deep autoencoder includes four convolutional layers. This allowed us to analyze how the complexity of the Task-Decoder affects the overall performance.

We trained all models using the Adam optimizer with learning rates set to $1 \times 10^{-4}$ for the encoder and classifier, and $1 \times 10^{-6}$ for the Task-Decoder. Each model was trained over 50,000 episodes using 5-way 1-shot and 5-way 5-shot scenarios. The query size was maintained at 15 for both training and evaluation phases. To enhance model robustness, we applied standard data augmentation techniques, including random cropping, horizontal flipping, and color jittering.

Furthermore, we investigated the effect of using multiple autoencoders per task by associating different autoencoders with specific layers of the encoder network. However, we found that using a single autoencoder provided a good trade-off between performance improvement and computational overhead. 

\begin{figure}
  
  \centering
   \includegraphics[width=1\linewidth]{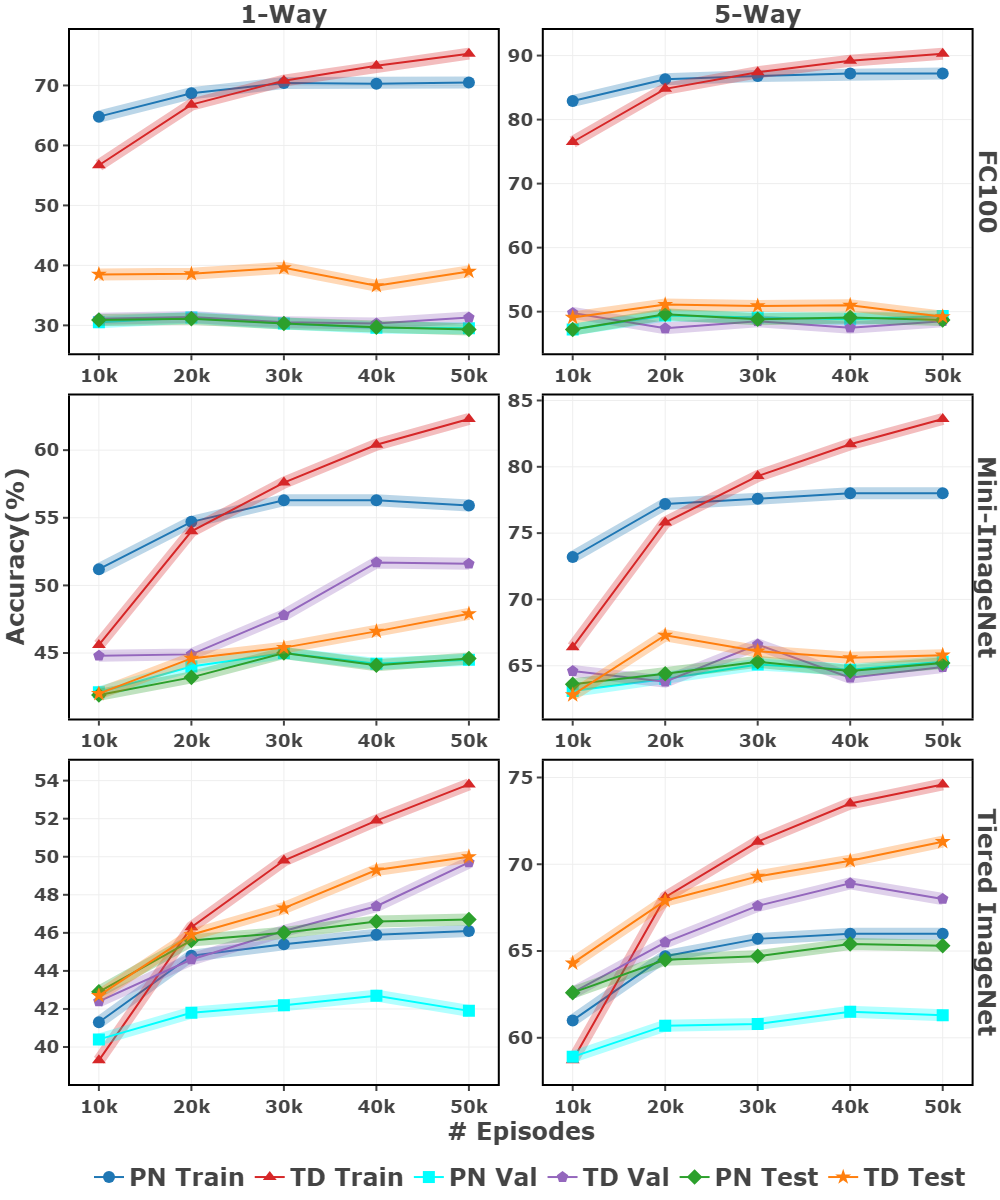}
   \caption{Accuracy curves for Prototypical Networks (PN) and Task-Decoder (TD) across different datasets, plotted against the number of episodes.}
   \label{fig:5}
\end{figure}
\begin{figure}
  
  \centering
   \includegraphics[width=1\linewidth]{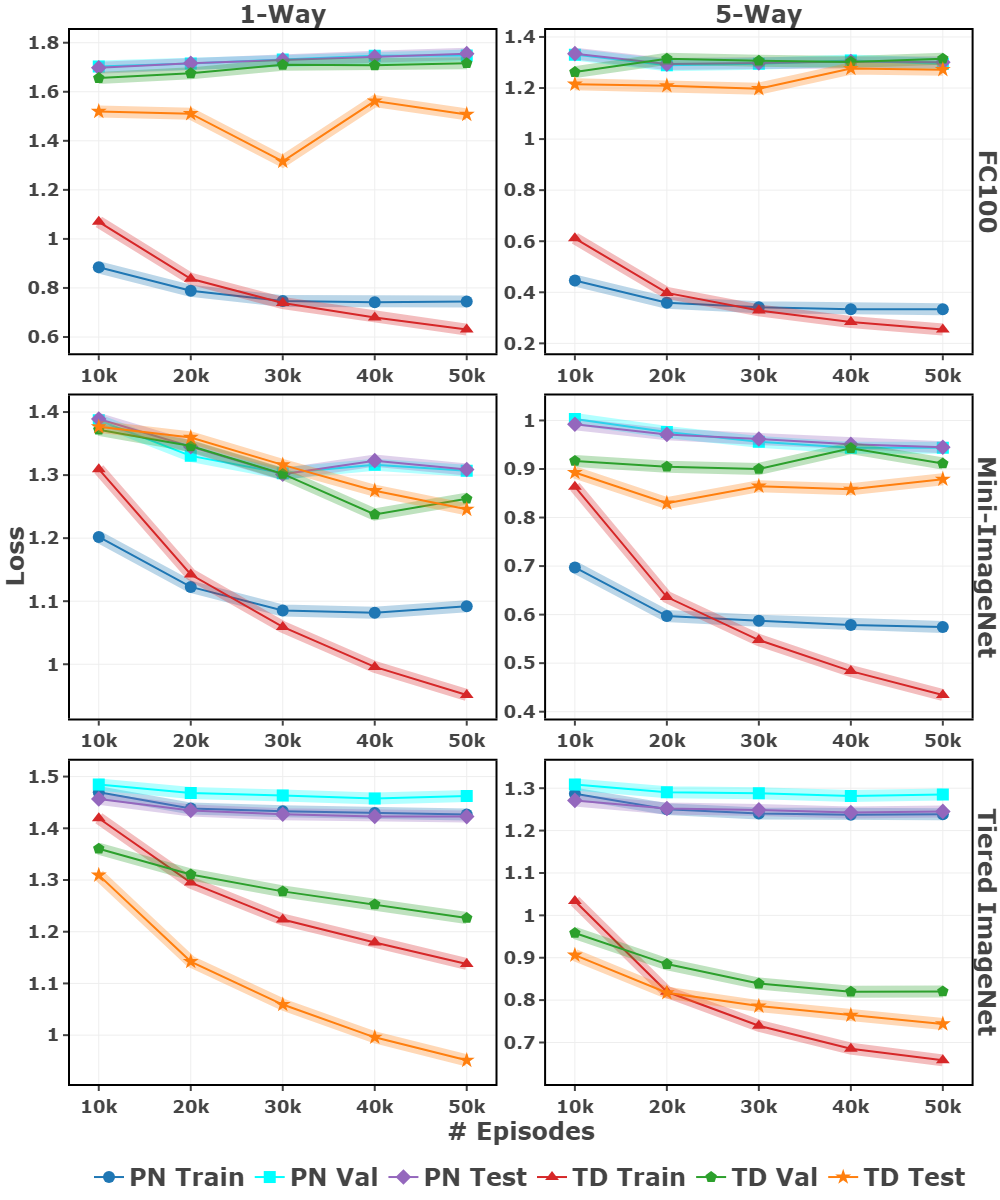}
   \caption{Loss curves for Prototypical Networks (PN) and Task-Decoder (TD) across different datasets, plotted against the number of episodes.}
   \label{fig:4}
\end{figure}
\begin{table*}[!htbp]
\centering
\resizebox*{!}{0.35\linewidth}{\begin{tabular}{ll|cccc}
                                               &                    & \multicolumn{2}{c}{MiniImageNet}                                        & \multicolumn{2}{c}{TieredImageNet}                                          \\
Method                                         & Backbone           & \multicolumn{1}{l}{5-way 1-shot(\%)} & \multicolumn{1}{l}{5-way 5-shot(\%)} & \multicolumn{1}{l}{5-way 1-shot(\%)} & \multicolumn{1}{l}{5-way 5-shot(\%)} \\
\hline & & \\
PN                           & ResNet18           & 45 ± 0.0                         & 65.2 ± 0.0                           & 46.6 ± 0.0                           & 65.4 ± 0.0                           \\
\textbf{PN(TD)}  & ResNet18           & \textbf{47.9 ± 0.0}              & \textbf{66.1 ± 0.0}                  & \textbf{50 ± 0.0}                    & \textbf{71.3 ± 0.0}                  \\
MAML                                           & CNN                & 47.54                        & 62.23                            & -                                    & -                                    \\
\textbf{MAML (TD)}                   & Shallow CNN Decoder                & \textbf{47.81}               & \textbf{62.52}                   & -                                    & -                                    \\
\textbf{MAML (TD)}                   & CNN & \textbf{47.67}               & \textbf{63.01}                   & -                                    & -                                    \\
Relation Networks\cite{relation}      & 64-96-128-256      & 50.44 ± 0.82                     & 65.32 ± 0.70                         & 54.48 ± 0.93                         & 71.32 ± 0.78                         \\
R2D2\cite{bertinetto2018meta}                   & 96-192-384-512     & 51.2 ± 0.6                       & 68.8 ± 0.1                           & -                                    & -                                    \\
Transductive Prop Nets\cite{liu2018learning} & 96-192-384-512     & 55.51 ± 0.86                     & 69.86 ± 0.65                         & 59.91 ± 0.94                         & 73.3 ± 0.75                          \\
SNAIL\cite{mishra2017simple}                   & ResNet-12          & 55.71 ± 0.99                     & 68.88 ± 0.92                         & -                                    & -                                    \\
AdaResNet\cite{munkhdalai2018rapid}              & ResNet-12          & 56.88 ± 0.62                     & 71.94 ± 0.57                         & -                                    & -                                    \\
TADAM \cite{oreshkin2018tadam2}                 & ResNet-12          & 58.5 ± 0.30                      & 76.7 ± 0.30                          & -                                    & -                                    \\
MetaOptNet                                     & ResNet-12          & 61.23 ± 0.29                     & 77.14 ± 0.21                         & 63.10 ± 0.34                         & 78.58 ± 0.27                         \\
\textbf{MetaOptNet (TD)}             & ResNet-12          & \textbf{61.35 ± 0.29}            & \textbf{77.3 ± 0.22}                 & \textbf{63.20 ± 0.34}                & \textbf{78.66 ± 0.27}\\
{\color{black}P$>$M$>$F}                                     & \color{black}IN1K,ViT-base          & \color{black}\textbf{72.80 ± 0.0}                     & \color{black}\textbf{90.70 ± 0.0}                         & -                         & -                        \\
{\color{black}\textbf{P$>$M$>$F(TD)}}             & \color{black}IN1K,ViT-base          & \color{black}\textbf{74.10 ± 0.0}            & \color{black}\textbf{92.30 ± 0.0}                 & -                & -\\
\hline

\end{tabular}}
\caption{Comparison of few-shot classification accuracies (\%) with 95\% confidence intervals on MiniImageNet and TieredImageNet datasets. Results are reported for 5-way 1-shot and 5-way 5-shot. `PN' and `TD' refer to Prototypical Networks and Task-Decoder, respectively.}
\label{table:1}
\end{table*}
\begin{table}[!htpb]
\centering
\resizebox*{!}{0.35\linewidth}{\begin{tabular}{l|cc}
                                              &  \multicolumn{2}{c}{FC100}                                                   \\
Method                                         & \multicolumn{1}{l}{5-way 1-shot(\%)} & \multicolumn{1}{l}{5-way 5-shot(\%)} \\
\hline & & \\
PN                            & 31.1 ± 0.0                           & 49.6 ± 0.0                           \\
\textbf{PN (TD)}  & \textbf{39 ± 0.0}                    & \textbf{51.1 ± 0.0}                  \\
TADAM\cite{oreshkin2018tadam2}                  & 40.1 ± 0.4                           & 56.1 ± 0.4                           \\
MetaOptNet                                     & 40.19 ± 0.25                         & 55.04 ± 0.26                         \\
\textbf{MetaOptNet (TD)}             & \textbf{40.59 ± 0.26}                & \textbf{55.38 ± 0.25}                \\
\hline
\end{tabular}}
\caption{Comparison of few-shot classification accuracies (\%) with 95\% confidence intervals on the FC100 dataset \cite{oreshkin2018tadam2}. Results are reported for 5-way 1-shot and 5-way 5-shot settings. `PN' and `TD' refer to Prototypical Networks and Task-Decoder, respectively.}
\label{table:2}
\end{table}
\subsection{Performance Evaluation}
{\color{black} 
Our experimental results demonstrate that the Meta-Task framework consistently outperforms baseline models across all datasets and settings, achieving higher accuracy with significantly fewer training episodes. For example, as shown in Figure~\ref{fig:5}, TD enables higher test accuracy within 20k episodes, whereas PN Test fails to achieve the same even after 50k episodes. Similarly, in Table~\ref{table:1}, TD P$>$M$>$F reaches 92.30\% accuracy in 3k episodes compared to P$>$M$>$F's 90.70\% after 4k episodes.

These results highlight the framework's efficiency in leveraging a small number of additional parameters to enhance generalization while reducing computational overhead. By introducing Meta-Tasks, the learning process is effectively regulated, eliminating the need for extensive parameter tuning or large-scale episodic training. This makes the Meta-Task framework both lightweight and practical for real-world applications. Further details on are provided in Table~\ref{table:1} and Table~\ref{table:2}.
}

\subsubsection*{Prototypical Network Results}
Table~\ref{table:1} presents our experimental results on MiniImageNet and TieredImageNet, where our proposed method, the Task-Decoder, demonstrates significant performance improvements over the baseline Prototypical Networks. Specifically, on TieredImageNet in the 5-way 5-shot setting, our approach achieved an increase from 65.4\% to 71.3\% in test set accuracy, reflecting a substantial 5.9\% absolute improvement. This gain indicates superior meta-generalization capabilities of our method.

In the 5-way 1-shot scenario, the Task-Decoder improved accuracy from 46.6\% to 50\% on TieredImageNet, highlighting its efficacy in scenarios with extremely limited data. These consistent improvements across different datasets and settings demonstrate the robustness and effectiveness of our approach.
\subsubsection*{MAML Results}
Table~\ref{table:1} also showcases the performance of the {Task-Decoder} integrated with MAML. Our method outperforms the baseline MAML by an average of 0.48\% in the 5-way 1-shot setting and 0.78\% in the 5-way 5-shot setting on MiniImageNet. Although the absolute improvements are modest, they demonstrate the effectiveness of the {Task-Decoder} in enhancing the adaptability of MAML.

To examine the impact of the {Task-Decoder's} architecture ($\phi$) on generalization, we trained MAML using two different{Task-Decoders}: a shallow decoder with 75K parameters and a deeper decoder with 250K parameters. Interestingly, the shallow decoder excelled in the 5-way 1-shot task, suggesting that deeper models are not always superior when data is scarce. In contrast, the deeper decoder performed better with more shots, highlighting the importance of model depth in specific contexts. This indicates that the Task-Decoder and its Meta-Task extensions can be effective even with a shallow network, which is beneficial in resource-constrained environments.

It's important to note that MAML requires the computation of second derivatives during optimization. However, to remain consistent with our claim of a method-agnostic framework, we chose to use a first-order approximation for the Task-Decoder's optimization. This decision avoids the computational overhead associated with second-order derivatives but may introduce approximation errors. Another limitation may stem from the inherent difficulties of optimizing the learning rate in MAML, which is complicated by the need for careful tuning due to its sensitivity to hyperparameters. While using the same learning rate as the Prototypical Networks simplifies comparisons, it likely constrained MAML's overall performance. These factors suggest that further optimization and fine-tuning could enhance the benefits of integrating the Task-Decoder with MAML.

\subsubsection*{MetaOptNet Results}
The results presented in Table~\ref{table:1} indicate that integrating the Task-Decoder with MetaOptNet \cite{lee2019meta} leads to modest but consistent performance improvements, with an average accuracy increase of 0.12\% in the 5-way 1-shot setting and 0.16\% in the 5-way 5-shot setting on MiniImageNet. While these gains are not as pronounced as those observed with other methods, they underscore the potential of the Task-Decoder to enhance performance even in well-optimized frameworks like MetaOptNet.

We attribute the smaller improvements to MetaOptNet's distinctive training setup, where variations in batch size and the number of shots significantly impact model performance. Due to hardware limitations, we reduced the batch size and the number of training shots, which may have constrained the performance gains. These findings suggest that further hyperparameter optimization, such as adjusting batch size or learning rate, could yield more substantial improvements.

{\color{black} \subsubsection*{P$>$M$>$F Results}
Due to the pre-trained nature of the encoder, we implemented a custom scheduler to align TD's optimization. Additionally, hardware constraints necessitated deviations from the original hyperparameters, such as reducing image sizes and query shots, which may have influenced the observed improvements. Despite these constraints, TD achieved a notable 1.45\% performance increase, demonstrating its flexibility and adaptability to transformer-based methods under constrained conditions.}

\subsection{Regularization Improvements}
Figure~\ref{fig:5} illustrates that our method achieves higher accuracy more quickly during all training, validation, and testing phases compared to the baseline, which demonstrates faster convergence. Additionally, Figure~\ref{fig:4} illustrates that the loss associated with our method decreases more rapidly and consistently remains lower throughout the training process across all training, validation, and test sets. The Task-Decoder achieves a mean loss of 1.23 with a variance of 0.001, whereas the baseline has a mean loss of 1.3 and a variance of 0.0002. This indicates more efficient learning and improved generalization.

To address concerns regarding implementation specifics, we have provided detailed information about the hyperparameters, training times, and autoencoder architectures in the supplementary material. We also conducted ablation studies to analyze the effect of different autoencoder architectures and the use of multiple autoencoders per task. The results indicate that while deeper autoencoders can provide marginal performance gains, the computational overhead increases significantly. Therefore, a single, appropriately sized autoencoder strikes a good balance between performance and efficiency.

\section{Conclusion}\label{sec:conclusion}

{\color{black}
In this paper, we introduced the {\textbf{Meta-Task}} framework, a method-agnostic approach for regularization in few-shot learning (FSL) that leverages Meta-Tasks as auxiliary tasks with a set of parameters to mitigate overfitting. Theoretically, we demonstrated that our approach achieves global optima with strong generalization across tasks. Empirically, we validated its efficacy through the {\textbf{Task-Decoder}} Meta-Task, which significantly improved performance on MiniImageNet and TieredImageNet benchmarks, achieving faster convergence and reduced generalization error with minimal hyperparameter tuning. Overall, our findings highlight the potential of the {{Meta-Task}} framework to improve FSL models, suggesting that even minor adjustments can lead to significant benefits and encourage further exploration of Meta-Task-based regularization in broader contexts.}




{\small\bibliographystyle{ieeenat_fullname}\bibliography{main}}

\clearpage
\setcounter{page}{1}
\maketitlesupplementary

\section*{Supplementary Material}
This supplementary material provides additional details and results to support the claims made in the main paper. It includes extended experimental results and detailed descriptions of the models, hyperparameters, and datasets used.

\section{Experimental Details}
\label{sec:ablation}
We trained all models using the recommended hyperparameters for each method. For the Task-Decoder, we maintained consistent settings across experiments. Table~\ref{sup:1} lists the hyperparameters used for training the Prototypical Networks and their Task-Decoder counterparts.

\begin{table}[htbp]
\centering
\resizebox{0.85\columnwidth}{!}{%
\begin{tabular}{l|cc}
\hline
\textbf{Hyperparameter} & \textbf{5-shot} & \textbf{1-shot} \\ \hline
Epoch      & 5    & 5    \\ 
Train Num Episodes & 10,000         & 10,000         \\ 
Test Num Episodes  & 10,000         & 10,000         \\ 
Train Way          & 5             & 5             \\ 
Train Shot         & 5             & 1             \\ 
Train Query        & 15            & 15            \\ 
Test Way           & 5             & 5             \\ 
Test Shot          & 5             & 1             \\ 
Test Query         & 15            & 15            \\ 
Optimizer           & Adam          & Adam          \\ 
Task-Decoder Learning Rate     & $1 \times 10^{-6}$      & $1 \times 10^{-6}$      \\ 
Learning Rate                  & $1 \times 10^{-4}$      & $1 \times 10^{-4}$      \\ \hline
\end{tabular}%
}
\caption{Hyperparameters used to train Prototypical Networks for 5-shot and 1-shot experiments.}
\label{sup:1}
\end{table}

Due to hardware constraints, we adjusted certain parameters for MetaOptNet, such as reducing the batch size from 8 to 6 and decreasing the number of training shots from 15 to 10. These adjustments were necessary to fit the models into GPU memory but may have affected performance.

{\color{black} Due to hardware constraints, we adjusted certain parameters for P$>$M$>$F, such as reducing the image size from 128 to 92 and decreasing the number of training shots from 15 to 10. These adjustments were necessary to fit the models into GPU memory but may have affected performance.}

\section{{\color{black}{Additional Meta-Tasks}}}
{\color{black} To further demonstrate the convergence of Meta-Tasks, we designed simple neural layers and implemented basic Meta-Tasks to compare their impact on performance. As shown in Tables~\ref{sup:6} and \ref{sup:7}, even these simple Meta-Tasks provide better regularization and improve performance. However, designing more complex Meta-Tasks requires careful attention and hyperparameter tuning to be effective. For instance, the GAN PN Meta-Task in Table\ref{sup:6} negatively impacted training performance, while all other Meta-Tasks led to improvements. Moreover, Table~\ref{sup:7} highlights that while not all Meta-Tasks yield dramatic performance gains, they generally contribute to performance improvements.}

\begin{table}[h]
\centering
\resizebox{0.95\columnwidth}{!}{%
\begin{tabular}{ll|c}
\hline
\color{black}\textbf{Meta-Task}  & \color{black}\textbf{Meta-Task Loss}  & \color{black}\textbf{Accuracy(\%)}        \\ \hline
\color{black}Vanilla             & \color{black}-                        & \color{black}64.70                 \\ 
\color{black}Task-Decoder (TD)   & \color{black}Reconstruction Loss      & \color{black}\textbf{67.30}         \\ 
\color{black}RL PN               & \color{black}Reward Function          & \color{black}66.88                  \\ 
\color{black}Rotating PN         & \color{black}Cross-Entropy            & \color{black}66.06                  \\ 
\color{black}MIXWAYS PN          & \color{black}Cross-Entropy            & \color{black}67.19                  \\ 
\color{black}GAN PN              & \color{black}Cross-Entropy            & \color{black}46.10                  \\ \hline
\end{tabular}%
}
\caption{\color{black}Performance comparison of different Meta-Tasks with vanilla Prototypical Networks in a 5-Way 5-Shot 5-Query for 50,000 episodes.}
\label{sup:6}
\end{table}

\begin{table}[]
\centering
\resizebox{0.95\columnwidth}{!}{%
\begin{tabular}{ll|c}
\hline
\color{black}\textbf{Meta-Task}       & \color{black}\textbf{Meta-Task Loss}      & \color{black}\textbf{Accuracy(\%)}         \\ \hline
\color{black}Vanilla          & \color{black}{-}                   & \color{black}24.00          \\ 
\color{black}TD               & \color{black}Reconstruction Loss & \color{black}\textbf{36.20} \\ 
\color{black}ArcFace          & \color{black}Cosine Distance     & \color{black}33.80          \\ 
\color{black}ArcMarginProduct & \color{black}Cosisne Distance     & \color{black}32.20          \\ 
\color{black}AddMarginProduct & \color{black}Cosine Distance     & \color{black}20.60          \\ 
\color{black}SphereProduct    & \color{black}Cosine Distance     & \color{black}23.80          \\ \hline
\end{tabular}%
}
\caption{\color{black}Performance comparison of different Meta-Tasks with vanilla Prototypical Networks in a 5-Way 5-Shot 5-Query for 500 episodes.}
\label{sup:7}
\end{table}

\section{\color{black}{Effect of Task Decoder of feature space}}
\begin{figure*}
  \centering
   \includegraphics[width=\linewidth]{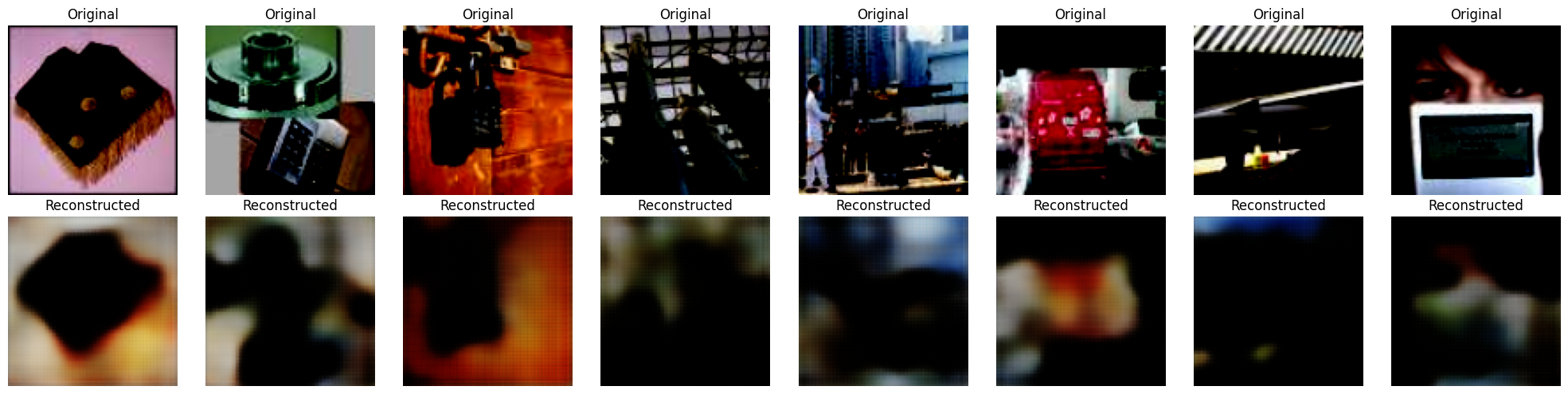} 
   \includegraphics[width=\linewidth]{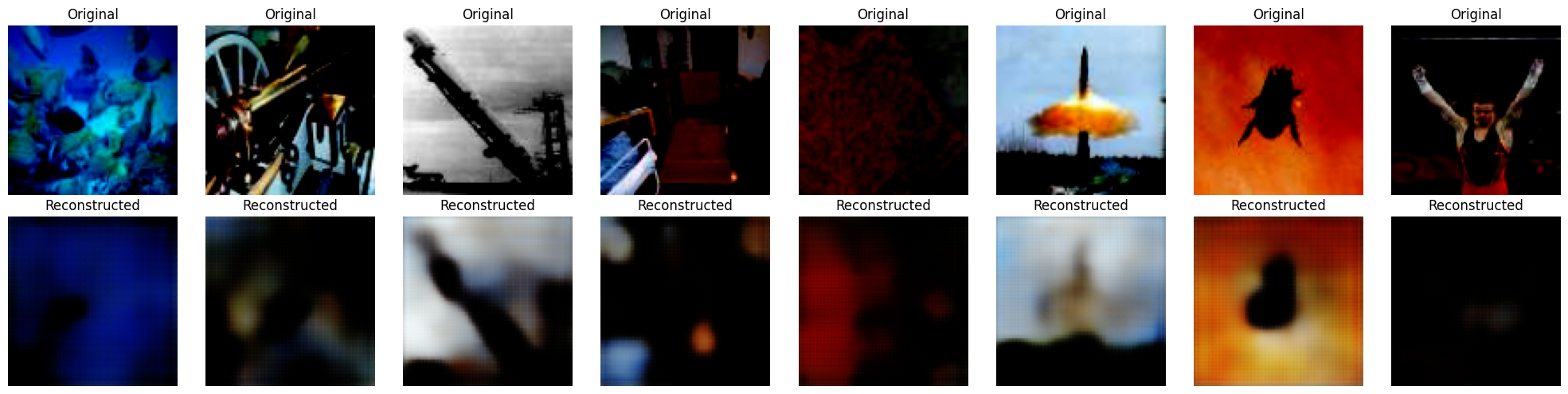}
   \includegraphics[width=\linewidth]{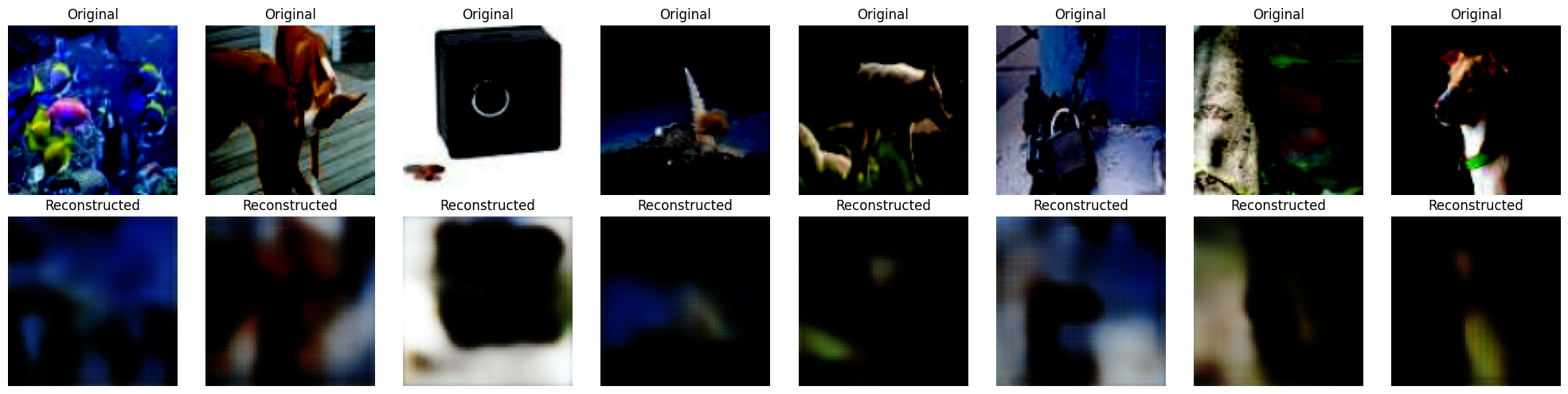}
    \caption{\color{black}The reconstruction results of the Prototypical Network integrated with the Task-Decoder, trained on TieredImageNet, are evaluated using MiniImageNet images.}
   \label{sup_fig:1}
\end{figure*}

\begin{figure*}
  \centering
\includegraphics[width=\linewidth]{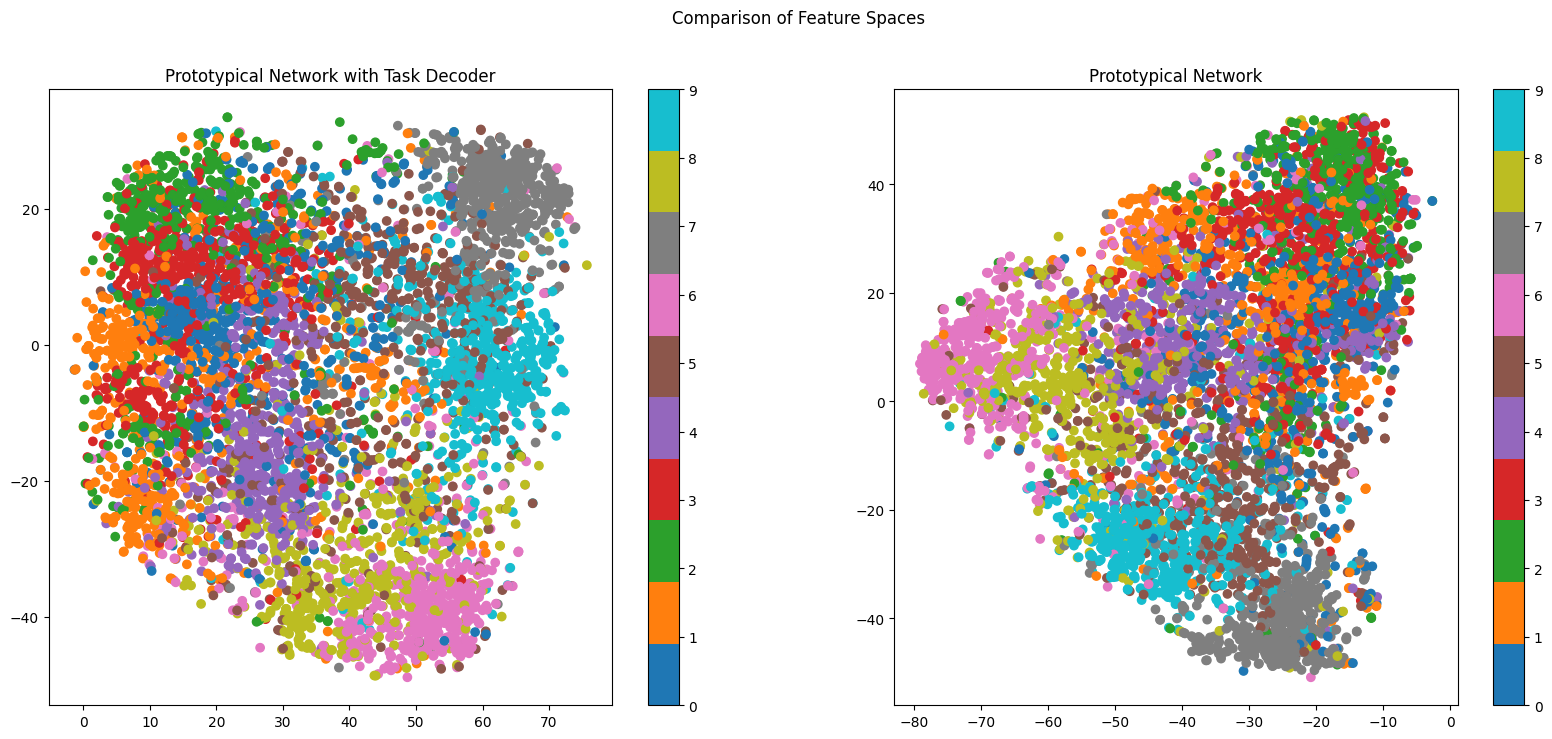} 
    \caption{\color{black}illustrates the t-SNE projections of the first 10 classes of MiniImageNet for Prototypical Networks with (left) and without (right) the Task-Decoder. The integration of the Task-Decoder demonstrates enhanced class separability, indicating its effectiveness in improving feature representation.}
   \label{sup_fig:2}
\end{figure*}

\begin{figure*}
  \centering
   \includegraphics[width=\linewidth]{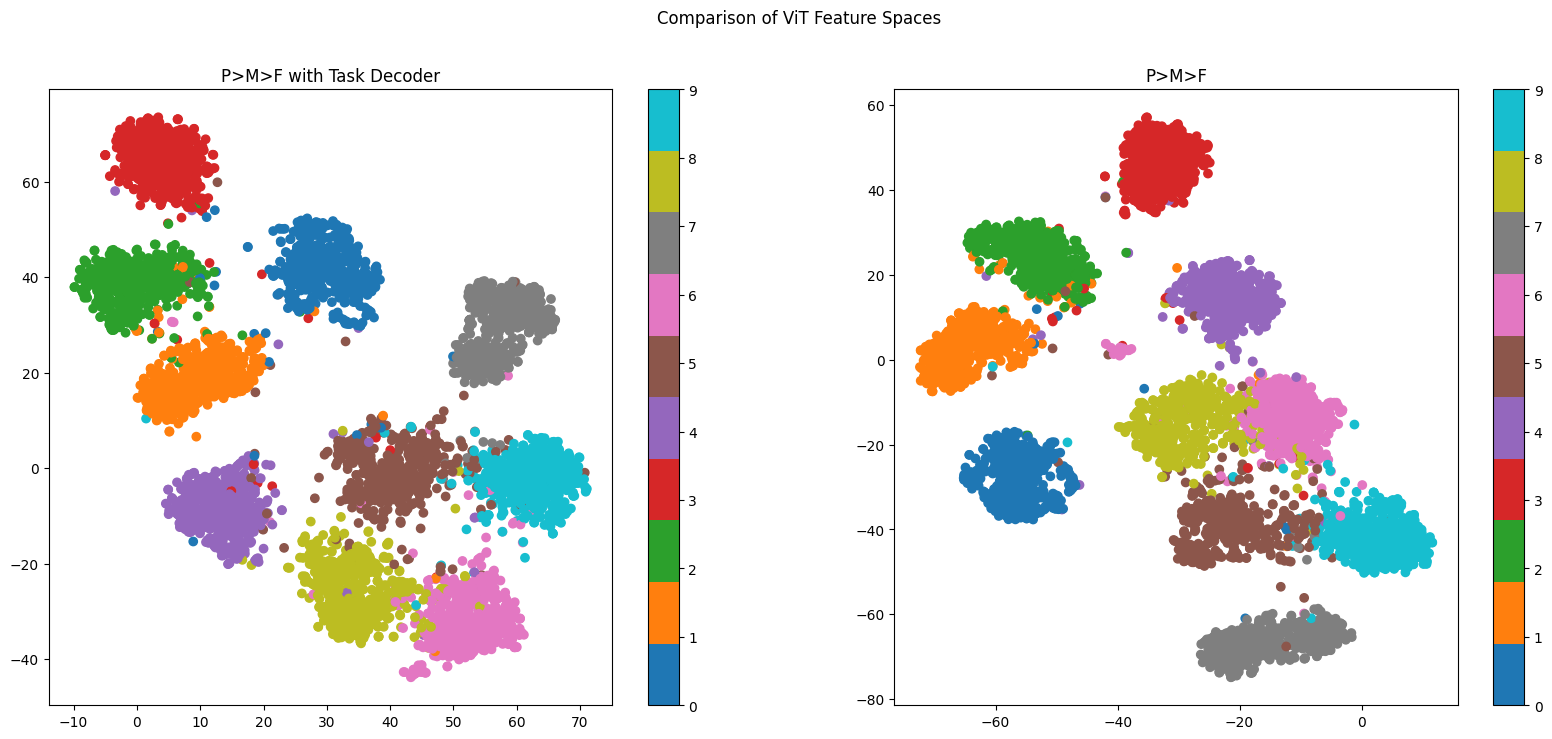} 
    \caption{\color{black}illustrates the t-SNE projections of the first 10 classes of MiniImageNet for P$>$M$>$F with (left) and without (right) the Task-Decoder. The integration of the Task-Decoder demonstrates enhanced class separability, indicating its effectiveness in improving feature representation.}
   \label{sup_fig:3}
\end{figure*}

\begin{figure*}
  \centering
   \includegraphics[width=0.8\linewidth]{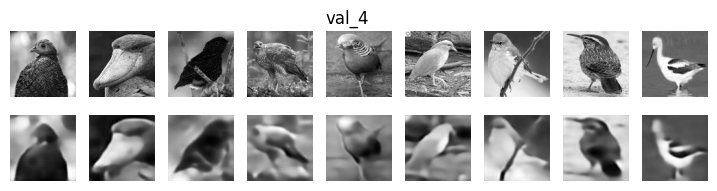} 
   \includegraphics[width=0.8\linewidth]{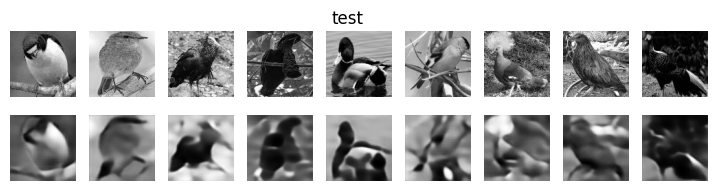}
    \caption{\color{black}Reconstruction results ResNet Task Decoder before few-shot training.}
   \label{sup_fig:4}
\end{figure*}

\begin{figure*}
  \centering
   \includegraphics[width=\linewidth]{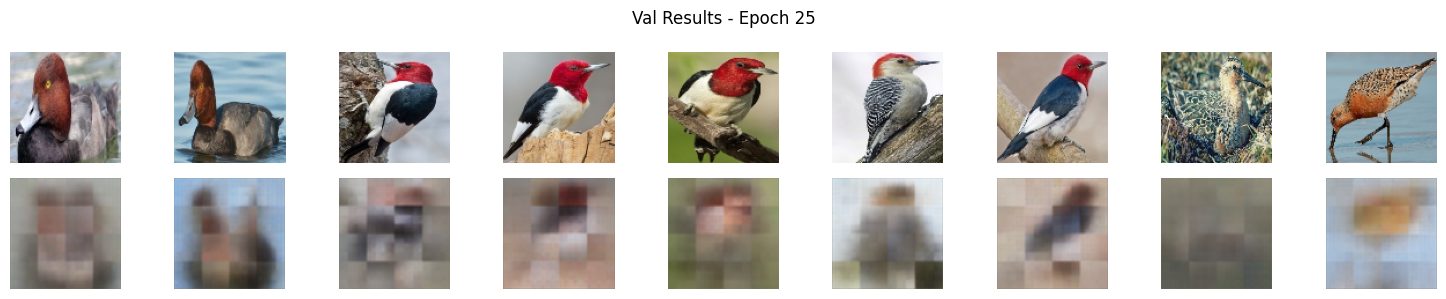} 
   \includegraphics[width=\linewidth]{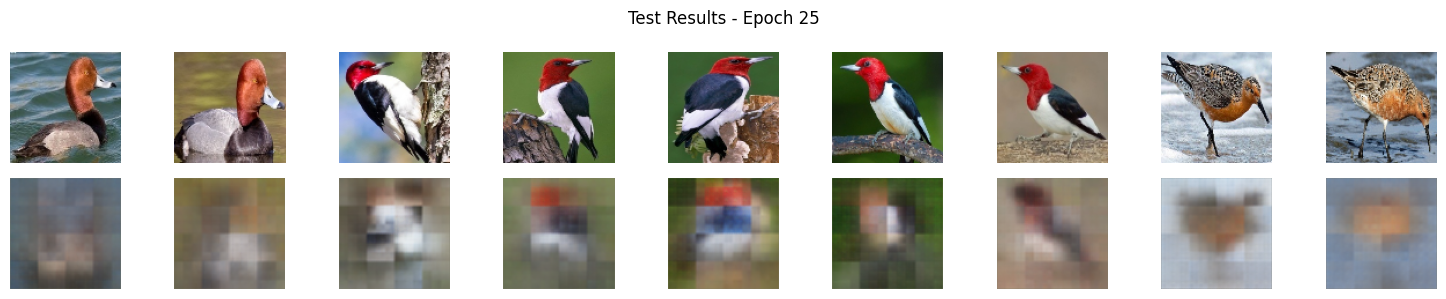}
    \caption{\color{black}Reconstruction results Task Decoder before few-shot training.}
   \label{sup_fig:5}
\end{figure*}

\begin{figure*}
  \centering
   \includegraphics[width=\linewidth]{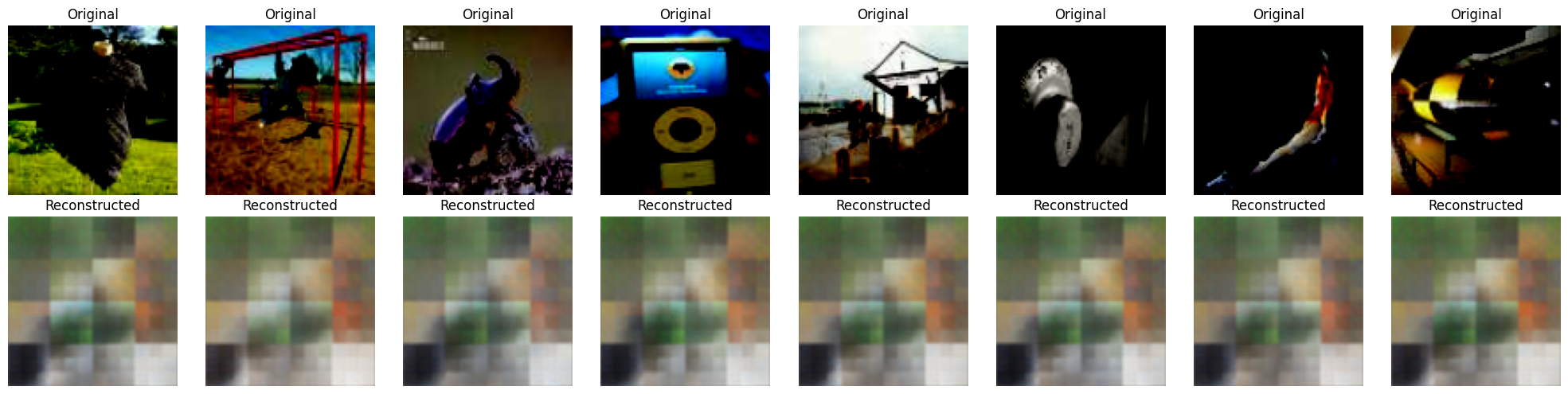} 
    \caption{\color{black}Reconstruction results of P$>$M$>$F with ViT Task Decoder trained on MiniImageNet images.}
   \label{sup_fig:6}
\end{figure*}

{\color{black} In this section, we analyze the effect of few-shot learning on the feature space and the Task-Decoder. Figure~\ref{sup_fig:4} shows the image reconstructions before few-shot learning, while Figure~\ref{sup_fig:1} shows the reconstructions after. Similarly, Figure~\ref{sup_fig:5} presents the reconstructions before few-shot learning, and Figure~\ref{sup_fig:6} depicts them after. Interestingly, the performance of the Task-Decoder changes following few-shot learning. Although the exact reasons for this behavior are unclear, several factors may contribute to this effect, including the imbalance in the number of parameters between the encoder (85M) and the decoder (3M), insufficient training of the decoder (only 25 episodes with approximately 5,000 images), or other underlying issues.

Additionally, Figure~\ref{sup_fig:2} and Figure~\ref{sup_fig:3} display the t-SNE projections of the first 10 classes of MiniImageNet with and without the Task-Decoder. Both visualizations exhibit improved clustering with the Task-Decoder, indicating better feature space representation. This suggests that the Task-Decoder enhances the performance of Prototypical Networks while improving generalization and regularization capabilities.}

\section{Extended Results: Prototypical Networks}

We evaluated test accuracies using standard classification metrics and computed the F1-score to compare the performance of the two models. Tables~\ref{sup:2}, ~\ref{sup:3}, ~\ref{sup:8}, ~\ref{sup:9} present a comparison between Prototypical Networks and their Task-Decoder counterparts, demonstrating that the Task-Decoder exhibits significantly superior adaptation-generalization compared to Prototypical Networks. The F1-scores further indicate superior meta-generalization to novel tasks by the Task-Decoder.

\begin{table*}[htbp]
\centering
\resizebox{1.3\columnwidth}{!}{%
\begin{tabular}{ccccc}
\hline
\textbf{Class} & \textbf{Support} & \textbf{Precision (\%)} & \textbf{Recall (\%)} & \textbf{F1-score (\%)} \\
\hline
0  & 46635  & 36.64 & 29.59 & 32.74 \\
1  & 44925  & 51.78 & 35.31 & 41.99 \\
2  & 37020  & 42.86 & 39.67 & 41.20 \\
3  & 30330  & 46.58 & 46.50 & 46.54 \\
4  & 44670  & 34.14 & 34.82 & 34.48 \\
5  & 29460  & 31.27 & 46.75 & 37.47 \\
6  & 36465  & 38.62 & 68.32 & 49.35 \\
7  & 44640  & 41.56 & 32.46 & 36.45 \\
8  & 22770  & 36.59 & 27.04 & 31.10 \\
9  & 14670  & 69.57 & 93.33 & 79.72 \\
10 & 36945  & 34.01 & 34.48 & 34.24 \\
11 & 38595  & 32.40 & 31.89 & 32.14 \\
12 & 37125  & 38.08 & 32.01 & 34.78 \\
13 & 59445  & 30.77 & 28.49 & 29.59 \\
14 & 46245  & 43.04 & 46.19 & 44.56 \\
15 & 44925  & 46.80 & 58.68 & 52.07 \\
17 & 67725  & 52.55 & 58.48 & 55.35 \\
18 & 29100  & 49.93 & 28.34 & 36.15 \\
19 & 38310  & 62.36 & 52.30 & 56.89 \\
\hline
\textbf{Accuracy (\%)}       & \multicolumn{4}{c}{42.90} \\
\textbf{Macro Average (\%)}  &        & 43.13 & 43.40 & 42.46 \\
\textbf{Weighted Average (\%)} &      & 42.45 & 42.20 & 41.64 \\
\hline
\end{tabular}%
}
\caption{5-way 1-shot classification report of Precision (\%), Recall (\%), and F1-score (\%) on MiniImageNet using \textbf{Prototypical Networks} after 50,000 episodes.}
\label{sup:2}
\end{table*}

\begin{table*}[htbp]
\centering
\resizebox{1.3\columnwidth}{!}{%
\begin{tabular}{ccccc}
\hline
\textbf{Class} & \textbf{Support} & \textbf{Precision (\%)} & \textbf{Recall (\%)} & \textbf{F1-score (\%)} \\
\hline
0  & 46635  & 40.08 & 31.99 & 35.58 \\
1  & 44925  & 57.97 & 47.64 & 52.30 \\
2  & 37020  & 53.45 & 43.70 & 48.09 \\
3  & 30330  & 59.32 & 52.89 & 55.92 \\
4  & 44670  & 40.26 & 37.58 & 38.87 \\
5  & 29460  & 45.57 & 58.67 & 51.30 \\
6  & 36465  & 45.14 & 73.41 & 55.90 \\
7  & 44640  & 45.68 & 41.21 & 43.33 \\
8  & 22770  & 53.30 & 36.17 & 43.09 \\
9  & 14670  & 66.96 & 96.80 & 79.16 \\
10 & 36945  & 39.39 & 38.42 & 38.90 \\
11 & 38595  & 41.17 & 39.37 & 40.25 \\
12 & 37125  & 44.74 & 39.99 & 42.23 \\
13 & 59445  & 36.32 & 36.61 & 36.46 \\
14 & 46245  & 45.68 & 46.86 & 46.26 \\
15 & 44925  & 53.66 & 54.31 & 53.98 \\
17 & 67725  & 53.84 & 67.44 & 59.88 \\
18 & 29100  & 45.04 & 31.58 & 37.13 \\
19 & 38310  & 56.64 & 56.94 & 56.79 \\
\hline
\textbf{Accuracy (\%)}       & \multicolumn{4}{c}{47.86} \\
\textbf{Macro Average (\%)}  &        & 48.64 & 49.03 & 48.18 \\
\textbf{Weighted Average (\%)} &      & 47.72 & 47.86 & 47.27 \\
\hline
\end{tabular}%
}
\caption{5-way 1-shot classification report of Precision (\%), Recall (\%), and F1-score (\%) on MiniImageNet using \textbf{Prototypical Networks with Task-Decoder} after 50,000 episodes.}
\label{sup:3}
\end{table*}

\begin{table*}[htbp]
\centering
\resizebox{1.3\columnwidth}{!}{%
\begin{tabular}{c|c|c|c|c}
\hline
\color{black}\textbf{Class} & \color{black}\textbf{Support} & \color{black}\textbf{Precision (\%)} & \color{black}\textbf{Recall (\%)} & \color{black}\textbf{F1-score (\%)} \\
\hline
\color{black}0  & \color{black}2370  & \color{black}100.00 & \color{black}98.06 & \color{black}99.02 \\
\color{black}1  & \color{black}3420  & \color{black}95.95  & \color{black}93.45 & \color{black}94.68 \\
\color{black}2  & \color{black}4155  & \color{black}92.43  & \color{black}98.72 & \color{black}95.47 \\
\color{black}3  & \color{black}6270  & \color{black}96.52  & \color{black}88.60 & \color{black}92.39 \\
\color{black}4  & \color{black}3825  & \color{black}95.86  & \color{black}92.05 & \color{black}93.92 \\
\color{black}5  & \color{black}810   & \color{black}100.00 & \color{black}86.67 & \color{black}92.86 \\
\color{black}6  & \color{black}4320  & \color{black}88.03  & \color{black}98.87 & \color{black}93.13 \\
\color{black}7  & \color{black}795   & \color{black}100.00 & \color{black}100.00 &\color{black} 100.00 \\
\color{black}8  & \color{black}3135  & \color{black}96.94  & \color{black}94.93 & \color{black}95.92 \\
\color{black}9  & \color{black}5535  & \color{black}89.10  & \color{black}86.97 & \color{black}88.02 \\
\color{black}10 & \color{black}2970  & \color{black}90.49  & \color{black}94.78 & \color{black}92.58 \\
\color{black}11 & \color{black}3720  & \color{black}93.28  & \color{black}94.81 & \color{black}94.04 \\
\color{black}12 & \color{black}3810  & \color{black}92.72  & \color{black}90.31 & \color{black}91.50 \\
\color{black}13 & \color{black}4710  & \color{black}95.02  & \color{black}87.90 & \color{black}91.32 \\
\color{black}14 & \color{black}3330  & \color{black}88.83  & \color{black}87.15 & \color{black}87.98 \\
\color{black}15 & \color{black}5145  & \color{black}93.43  & \color{black}96.40 & \color{black}94.89 \\
\color{black}16 & \color{black}4080  & \color{black}89.26  & \color{black}90.44 & \color{black}89.85 \\
\color{black}17 & \color{black}5010  & \color{black}95.29  & \color{black}93.71 & \color{black}94.50 \\
\color{black}18 & \color{black}3600  & \color{black}87.67  & \color{black}81.58 & \color{black}84.52 \\
\color{black}19 & \color{black}3990  & \color{black}81.23  & \color{black}95.89 & \color{black}87.95 \\
\hline
\color{black}\textbf{Accuracy (\%)}       & \multicolumn{4}{c|}{\color{black}92.25} \\
\color{black}\textbf{Macro Average (\%)}  & \color{black}75000   & \color{black}93.10  & \color{black}92.57 & \color{black}92.73 \\
\color{black}\textbf{Weighted Average (\%)} & \color{black}75000 & \color{black}92.44  & \color{black}92.25 & \color{black}92.25 \\
\hline
\end{tabular}%
}
\caption{\color{black}5-way 5-shot classification report of Precision (\%), Recall (\%), and F1-score (\%) on MiniImageNet using \textbf{P$>$M$>$F with Task Decoder} after 3,000 episodes.}
\label{sup:8}
\end{table*}

\begin{table*}[htbp]
\centering
\resizebox{1.3\columnwidth}{!}{%
\begin{tabular}{c|c|c|c|c}
\hline
\color{black}\textbf{Class} & \color{black}\textbf{Support} & \color{black}\textbf{Precision (\%)} & \color{black}\textbf{Recall (\%)} & \color{black}\textbf{F1-score (\%)} \\
\hline
\color{black}0  & \color{black}5700  & \color{black}98.35  & \color{black}96.18  & \color{black}97.25 \\
\color{black}1  & \color{black}5100  & \color{black}93.46  & \color{black}91.14  & \color{black}92.29 \\
\color{black}2  & \color{black}2460  & \color{black}97.81  & \color{black}96.18  & \color{black}96.99 \\
\color{black}3  & \color{black}2040  & \color{black}100.00 & \color{black}82.45  & \color{black}90.38 \\
\color{black}4  & \color{black}2685  & \color{black}95.23  & \color{black}89.16  & \color{black}92.09 \\
\color{black}5  & \color{black}4530  & \color{black}78.52  & \color{black}95.36  & \color{black}86.12 \\
\color{black}6  & \color{black}4965  & \color{black}95.35  & \color{black}95.91  & \color{black}95.63 \\
\color{black}7  & \color{black}1485  & \color{black}92.70  & \color{black}94.07  & \color{black}93.38 \\
\color{black}8  & \color{black}1635  & \color{black}100.00 & \color{black}63.55  & \color{black}77.71 \\
\color{black}9  & \color{black}3630  & \color{black}98.21  & \color{black}88.95  & \color{black}93.35 \\
\color{black}10 & \color{black}3105  & \color{black}92.10  & \color{black}93.88  & \color{black}92.98 \\
\color{black}11 & \color{black}2895  & \color{black}94.66  & \color{black}90.09  & \color{black}92.32 \\
\color{black}12 & \color{black}4785  & \color{black}92.77  & \color{black}92.54  & \color{black}92.66 \\
\color{black}13 & \color{black}4605  & \color{black}91.58  & \color{black}89.55  & \color{black}90.56 \\
\color{black}14 & \color{black}6090  & \color{black}85.71  & \color{black}89.36  & \color{black}87.50 \\
\color{black}15 & \color{black}2145  & \color{black}91.15  & \color{black}97.53  & \color{black}94.23 \\
\color{black}16 & \color{black}3900  & \color{black}88.29  & \color{black}94.74  & \color{black}91.40 \\
\color{black}17 & \color{black}4425  & \color{black}95.97  & \color{black}87.19  & \color{black}91.37 \\
\color{black}18 & \color{black}5040  & \color{black}80.83  & \color{black}79.48  & \color{black}80.15 \\
\color{black}19 & \color{black}3780  & \color{black}79.10  & \color{black}94.31  & \color{black}86.04 \\
\hline
\color{black}\textbf{Accuracy (\%)}       & \multicolumn{4}{c|}{\color{black}90.74} \\
\color{black}\textbf{Macro Average (\%)}  & \color{black}75000 & \color{black}92.09  & \color{black}90.08  & \color{black}90.72 \\
\color{black}\textbf{Weighted Average (\%)} & \color{black}75000 & \color{black}91.24  & \color{black}90.74  & \color{black}90.75 \\
\hline
\end{tabular}%
}
\caption{\color{black}5-way 5-shot classification report of Precision (\%), Recall (\%), and F1-score (\%) on MiniImageNet using \textbf{P$>$M$>$F} after 4,000 episodes.}
\label{sup:9}
\end{table*}

\section{Extended Results: MAML}

\subsection{Performance Comparison}

Table~\ref{sup:4} provides a detailed comparison of the performance of MAML and Task-Decoder models on the MiniImageNet dataset. The table shows results for both 5-way 1-shot and 5-way 5-shot classification tasks across training, validation, and test sets. Notably, the Task-Decoder with a deeper CNN architecture consistently outperforms the standard MAML across all metrics, particularly in test accuracy, indicating its superior ability to generalize to unseen data. These results highlight the effectiveness of the Task-Decoder in improving model performance on few-shot learning tasks.

\begin{table*}[htbp]
\centering
\resizebox{0.7\textwidth}{!}{%
\begin{tabular}{ll|ccc|ccc}
\hline
\multicolumn{2}{c|}{\textbf{Method}} & \multicolumn{3}{c|}{\textbf{5-way 1-shot (\%)}} & \multicolumn{3}{c}{\textbf{5-way 5-shot (\%)}} \\
\hline
\textbf{Model} & \textbf{Backbone} & \textbf{Train} & \textbf{Val} & \textbf{Test} & \textbf{Train} & \textbf{Val} & \textbf{Test} \\
\hline
MAML & CNN & 57.58 & 46.36 & 47.54 & 76.64 & 61.11 & 62.23 \\
\textbf{MAML (Task-Decoder)} & Shallow CNN & \textbf{58.10} & \textbf{45.24} & \textbf{47.81} & \textbf{76.82} & \textbf{61.04} & \textbf{62.52} \\
\textbf{MAML (Task-Decoder)} & CNN & \textbf{57.94} & \textbf{46.13} & \textbf{47.67} & \textbf{76.84} & \textbf{61.57} & \textbf{63.01} \\
\hline
\end{tabular}%
}
\caption{Comparison of few-shot classification accuracies (\%) on MiniImageNet using MAML and Task-Decoder models. Results are reported for 5-way 1-shot and 5-way 5-shot settings, including Train, Validation, and Test accuracies.}
\label{sup:4}
\end{table*}

\subsection{Computational Analysis}

To provide insights into the computational requirements of our approach, Table~\ref{sup:5} presents the training times for various model configurations. {\color{black}It is important to note that the reported times correspond to training for exactly 50,000 episodes. However, in practice, the Task-Decoder (TD) often converges in significantly fewer episodes, further highlighting its computational efficiency.}


\begin{table*}[htbp]
\centering
\resizebox{0.6\textwidth}{!}{%
\begin{tabular}{ll|cc}
\hline
\textbf{Model} & \textbf{Backbone} & \textbf{5-way 1-shot} & \textbf{5-way 5-shot} \\
\hline
MAML & CNN & 5.1h & 7.2h \\
\textbf{MAML (Task-Decoder)} & Shallow CNN & \textbf{5.6h} & \textbf{7.9h} \\
\textbf{MAML (Task-Decoder)} & CNN & \textbf{6.4h} & \textbf{8.4h} \\
\hline
\end{tabular}%
}
\caption{Comparison of training times (in hours) on MiniImageNet using MAML and Task-Decoder models for 5-way 1-shot and 5-way 5-shot experiments.}
\label{sup:5}
\end{table*}

The Task-Decoder variants show consistently superior performance across metrics, particularly in test accuracy, while maintaining reasonable computational requirements. The deeper CNN architecture demonstrates the best overall performance, though at the cost of increased training time.

\end{document}